\documentclass{article}

\usepackage[table]{xcolor}
\usepackage[nonatbib, final]{neurips_2022}

\usepackage{highlight}
\renewcommand{\opacity}{50}
\usepackage[utf8]{inputenc} % allow utf-8 input
\usepackage[T1]{fontenc}    % use 8-bit T1 fonts
\usepackage{hyperref}       % hyperlinks
\usepackage{url}            % simple URL typesetting
\usepackage{booktabs}       % professional-quality tables
\usepackage{amsfonts}       % blackboard math symbols
\usepackage{nicefrac}       % compact symbols for 1/2, etc.
\usepackage{microtype}      % microtypography
\usepackage{xcolor}         % colors
\usepackage{graphicx}
\usepackage[fleqn]{amsmath}
\usepackage{subcaption}
\usepackage[english]{babel}
\usepackage[square, numbers]{natbib}
\usepackage{dsfont}
\usepackage{hyperref}

\newcommand*{\affmark}[1][*]{\textsuperscript{#1}}

\title{Joint Embedding Predictive Architectures Focus on Slow Features}

\author{%
Vlad Sobal\affmark[1] \quad Jyothir S V\affmark[1] \quad  Siddhartha Jalagam\affmark[1] \quad Nicolas Carion\affmark[2] \quad 
\\ \textbf{Kyunghyun Cho\affmark[1, 3, 4] \quad Yann LeCun \affmark[1, 2]} 
\\ \affmark[1]New York University \quad \affmark[2]Meta AI \quad \affmark[3]Prescient Design, Genentech \quad \affmark[4]CIFAR Fellow \\ 
\texttt{\{us441, jyothir, scj9994, carion.nicolas, kyunghyun.cho\}@nyu.edu}  \\ \texttt{yann@cs.nyu.edu}
}

\begin{document}

\maketitle

\begin{abstract}
Many common methods for learning a world model for pixel-based environments use generative architectures trained with pixel-level reconstruction objectives.
Recently proposed Joint Embedding Predictive Architectures (JEPA) \citep{lecun2022path} offer a reconstruction-free alternative. In this work, we analyze performance 
of JEPA trained with VICReg and SimCLR objectives in the fully offline setting without access to rewards, and compare the results to the performance of the 
generative architecture. We test the methods in a simple environment with a moving dot with various background distractors, and probe learned representations for the dot's location.
We find that JEPA methods perform on par or better than reconstruction when distractor noise changes every time step, but fail when the noise is fixed. Furthermore, we provide a theoretical explanation for the poor performance of JEPA-based methods with fixed noise, highlighting an important limitation.

\end{abstract}

\section{Introduction}

\looseness=-1
Currently, the most common approach to learning world models is to use reconstruction objectives \citep{dreamer, dreamer_v2, planet, atari_video_fm}.
However, reconstruction objectives suffer from object-vanishing, as the objectives by design do not distinguish important objects in the scene \citep{dreaming}. The framework of Joint Embedding Predictive Architectures presented in \cite{lecun2022path} may offer a possible alternative to reconstruction-based objectives, as has been shown in \cite{sgi, spr, dreaming}.
In this paper, we implement\footnote[1]{Code is available at \href{https://github.com/vladisai/JEPA_SSL_NeurIPS_2022}{\texttt{https://github.com/vladisai/JEPA\_SSL\_NeurIPS\_2022}}} JEPA for learning from image and action sequences with VICReg \cite{vicreg} and SimCLR \cite{simclr} objectives (section \ref{sec:method} and code).
We then test JEPA, reconstruction-based, and inverse dynamics modeling methods by training on image sequences of one moving dot and probing the learned representations for dot location in the presence of various noise distractors (section \ref{sec:experiments}). We observe that JEPA methods can learn to ignore distractor noise that changes every time step, but fail when distractor noise is static.
% waste of space
\section{Method}
\label{sec:method}
\begin{figure}
  \begin{subfigure}[t]{0.49\linewidth}
     \centering
     \includegraphics[height=4.5cm]{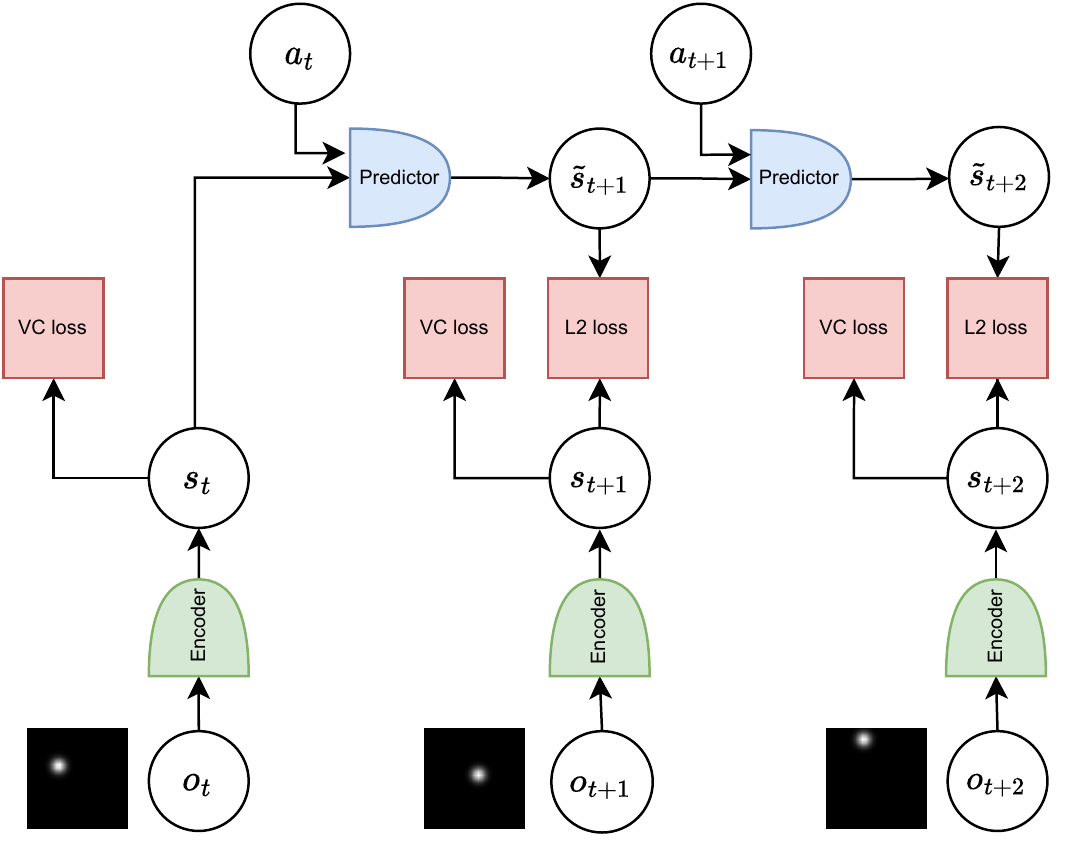}
     \caption{\textbf{VICReg-based architecture}}
     \label{fig:vicreg}
  \end{subfigure}
  \begin{subfigure}[t]{0.49\linewidth}
     \centering
     \includegraphics[height=4.5cm]{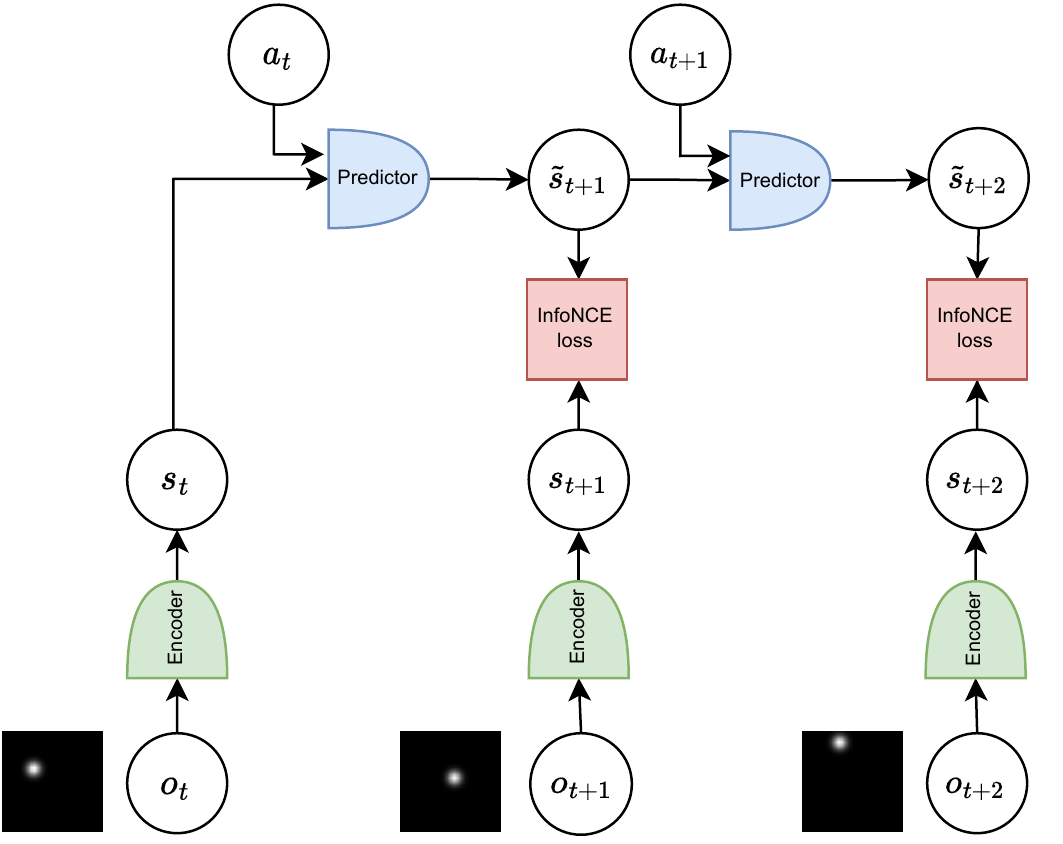}
     \caption{\textbf{SimCLR-based architecture}}
     \label{fig:simclr}
 \end{subfigure}
 \caption{\textbf{JEPA-based methods' training diagrams.} \textbf{(a)} JEPA with VICReg loss. VC here denotes variance and covariance losses from \cite{vicreg}. \textbf{(b)} JEPA with InfoNCE loss. }
 \vspace{-0.5cm}
\end{figure}

We focus on learning a world model from a fixed set of state-action sequences. We consider a Markov Decision Process (MDP) $M = (O, A, P, R)$. $O$ is a set of possible observations (in our case these are images), $A$ represents the set of possible actions, $P = Pr(o_t \vert o_{t-1}, a_{t-1})$ represents transition probabilities, and $R : O \times A \to \mathbb{R}$ is the reward function.
In the offline setting we focus on, we do not have access to the MDP directly, we only have a pre-recorded set of sequences of observations $o_t \in O$ and actions $a_t \in A$. 
Our goal is to use these offline sequences to learn an encoder that converts observations $o_t$ to $D$-dimensional representations $s_t$,  $g_\phi(o_t) = s_t, g_\phi : O \to \mathbb{R}^D$, and the forward model  
$f_\theta(s_t, a_t) = \tilde s_{t+1}, f_\theta : \mathbb{R}^D \times A \to \mathbb{R}^D$. 
During pre-training we do not have access to the evaluation task, therefore we aim to learn representations that capture in $D$-dimensional vectors as much information about the environment and its dynamics as possible. 

During both training and evaluation, given the initial observation $o_1$ and a sequence of actions of length $T$, $a_1 \dots a_T$, we encode the first observation $\tilde s_1 = g_\phi(o_1)$,
then auto-regressively apply the forward model $\tilde s_t = f_\theta(\tilde s_{t-1}, a_{t-1})$, obtaining representations for a sequence of length $T+1$: $\tilde s_1 \dots \tilde s_{T+1}$. During testing, we probe the representations using
a single linear layer that is trained, with frozen encoder and predictor, to recover some known properties of states $q(\tilde s_t) : \mathbb{R}^D \to \mathbb{R}^Q$, where $Q$ is the dimension of the target value.
This protocol is related to the one used in \cite{st-dim}, but we probe not only the encoder output, but also the predictor output. Fore more details on probing, see appendix \ref{ap:probing}. We compare multiple ways of training the encoder and predictor. In all the approaches, the gradients are propagated through time, and no image augmentations are used. Encoder and predictor architectures are fixed (see appendix \ref{ap:architectures}). We test the following methods:

\textbf{VICReg} (figure \ref{fig:vicreg})  We take inspiration from \cite{lecun2022path}, and adopt VICReg \citep{vicreg} objective for training a Joint Embedding Predictive Architecture (JEPA). We apply variance and covariance losses 
described in \cite{vicreg} to representations separately at each step, and apply the prediction loss to make the forward model 
output $\tilde s_t$ close to encoder output $s_t$. For a more detailed description, see appendix \ref{sec:ap_vicreg}.

\textbf{SimCLR} (figure \ref{fig:simclr})  In this case, 
% like in the above, 
we train JEPA, but instead of VICReg we utilize SimCLR objective \cite{simclr}. We apply InfoNCE loss \cite{cpc}, with the forward model output and the encoder output for the same time step as positive pairs. For a more detailed description, see appendix \ref{sec:ap_simclr}. 

\textbf{Reconstruction} Reconstruction approach introduces a decoder $d_\xi(\tilde s_t) = \tilde o_t$, and utilizes a reconstruction objective $\mathcal{L} = \frac{1}{T}\sum_{t=1}^{T} \Vert o_t - \tilde o_t \Vert_2^2$ to train the encoder and predictor. See appendix \ref{ap:rssm} for more details.

\textbf{Inverse Dynamics Modeling (IDM)}  We add a linear layer that, given the encoder's outputs at two consecutive
steps $g_\phi(o_t), g_\phi(o_{t+1})$, predicts $a_t$. The forward model is trained by predicting the encoder's output at the next time step. For more details, see appendix \ref{ap:idm}.

\textbf{Supervised} All components are trained end-to-end by propagating the error from the probing function
to both the encoder and decoder. This should give us the lower bound on probing error.

\textbf{Random} In this case, the probing is run with fixed random weights of the encoder and predictor.

\subsection{Spurious correlation}

VICReg and SimCLR JEPA methods may fall prey to a spurious correlation issue: the loss can be easily minimized by paying attention to noise that does not change with time, making the system ignore all other information. 
Intuitively, the objectives make the model focus on `slow features' \citep{slow_feature_analysis}. When
the slowest features in the input belong to fixed background noise, the representation will only contain the noise.
To demonstrate that, we show a trivial but plausible solution.
In the presence of normally distributed distractor noise that does not change with time, the model can extract features by directly copying the values of the noise from the input : $g_\phi(o_t) = s \sim \mathcal{N}(0, \sigma^2 I), \,s \in \mathbb{R}^D$. Since the noise is persistent through time, 
% we note that 
$g_\phi(o_t) = g_\phi(o_{t+1}) = s$. We assume that the forward model has converged to identity: $f_\theta(s, a) = s$. We denote a batch of encoded representations at step $t$ with a matrix $S_t \in \mathbb{R}^{N \times D}$ where $N$ is batch size; and batches of actions and observations with $A_t$ and $O_t$ respectively. Then the VICReg losses are:
\begin{align}
\mathcal{L}_\mathrm{prediction} &= \frac{1}{TN}\sum_{t=1}^{T}\sum_{i=1}^N\Vert f_\theta(S_{t, i}, A_{t,i}) - g_\phi(O_{t+1, i})\Vert_2^2  \nonumber \\
&= \frac{1}{TN}\sum_{t=1}^{T}\sum_{i=1}^N\Vert S_{t, i} - S_{t+1, i}\Vert_2^2 = \frac{1}{TN}\sum_{t=2}^{T+1}\sum_{i=1}^N\Vert S_{t, i} - S_{t, i}\Vert_2^2 = 0
\end{align}
\begin{align}
&\mathrm{Var}(s_t) = \frac{1}{N-1}\sum_{i=1}^N (s_i - \bar s)^2 = \sigma^2 & \text{As } s \sim \mathcal{N}(0, \sigma^2 I)
\end{align}
\begin{align}
&\mathcal{L}_\mathrm{variance} = \frac{1}{(T+1)D}\sum_{t=1}^{T+1}\sum_{j=1}^D \max \left(0, \gamma - \sqrt{\mathrm{Var}(S_{t, :, j}) + \epsilon}\right)  = 0 \label{eq:var} \\
&\mathcal{L}_\mathrm{covariance} =  \frac{1}{(T+1)(N-1)}\sum_{t=1}^{T+1}\sum_{i=1}^D \sum_{j=i+1}^D (S_t S_t^\top)_{i,j} = 0 & \label{eq:cov}
\end{align}

Equation \ref{eq:var} holds for large enough $\sigma$, \ref{eq:cov} holds because the noise variables are independent across episodes. The total sum of the loss components is then 0 for the described trivial solution. 

In SimCLR case, as shown by \citet{Wang_Isola_2022} in their theorem 1, the InfoNCE loss
is minimized in the infinite limit of negative samples if the positive pairs are perfectly aligned and the encoder output is perfectly uniform on the unit sphere. 
Both conditions are satisfied for the trivial solution described above, therefore SimCLR objective is also susceptible to fixed distractor noise problem.

\section{Experiments}
\label{sec:experiments}
\begin{figure}
  \begin{subfigure}[b]{0.47\linewidth}
     \centering
     \includegraphics[width=0.9\textwidth]{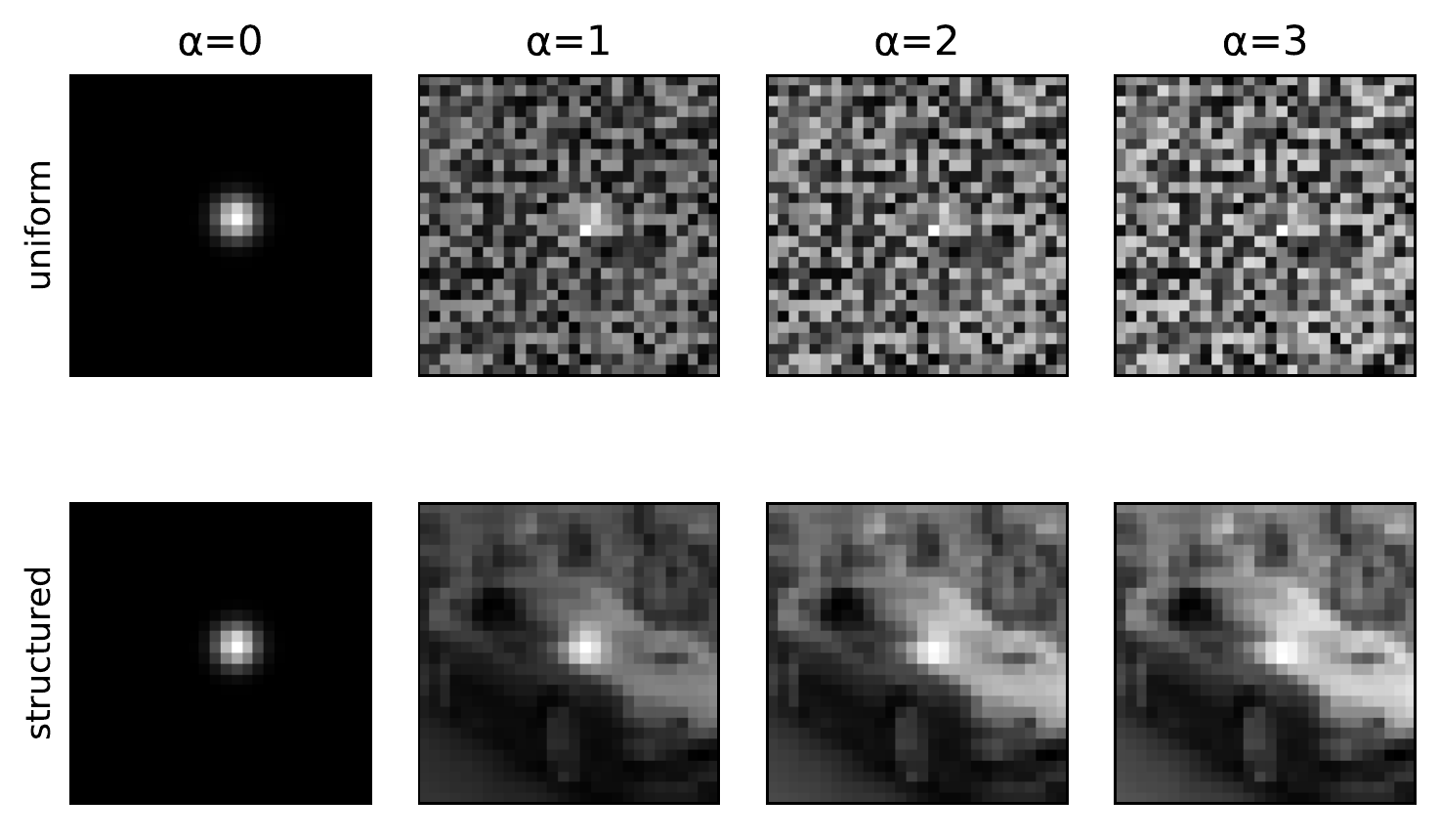}
     \caption{\textbf{Different noise levels}}
     \label{fig:static_dataset}
  \end{subfigure}
  \hfill
  \begin{subfigure}[b]{0.47\linewidth}
     \centering
     \includegraphics[width=0.9\textwidth]{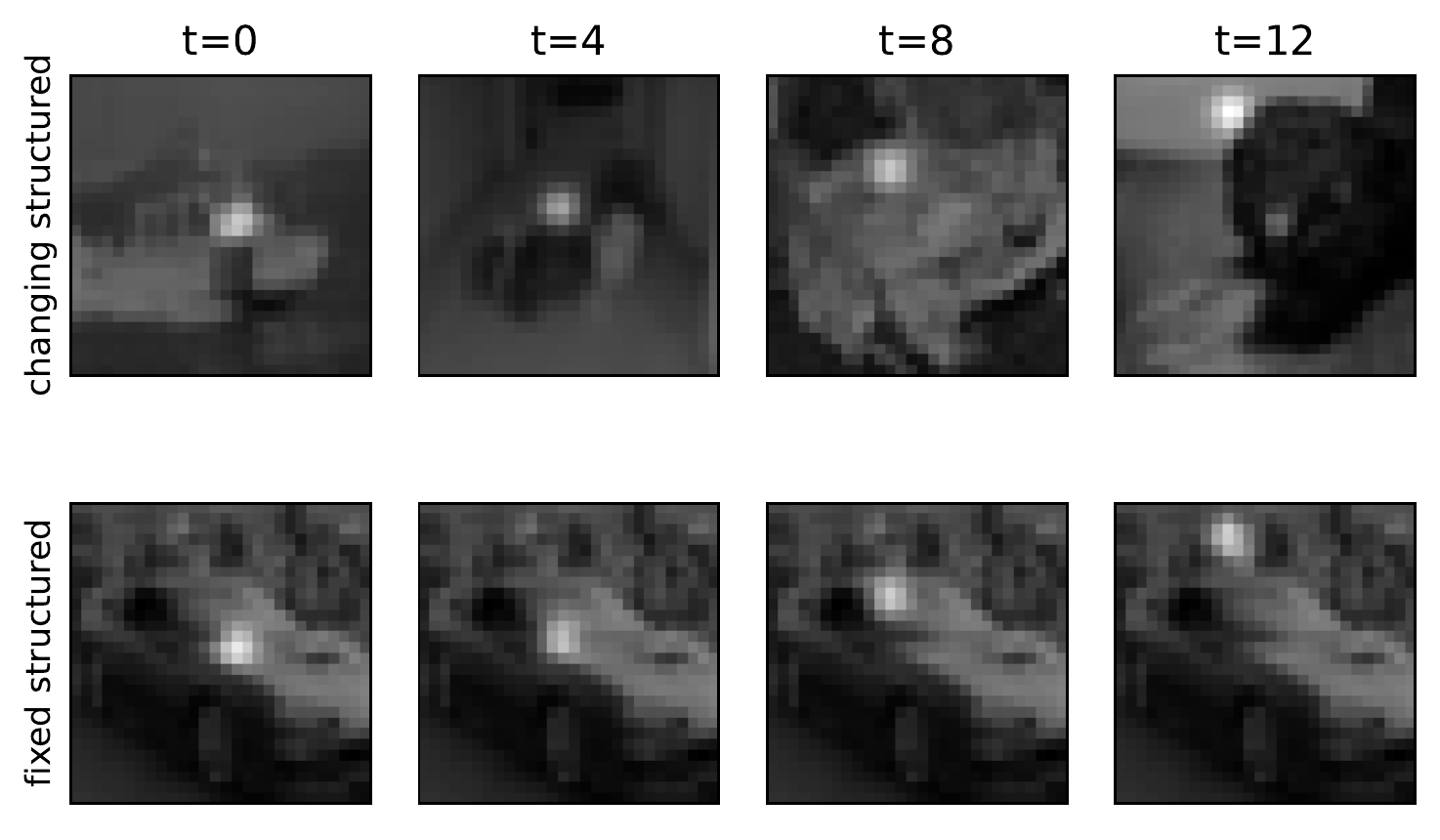}
     \caption{\textbf{Changing and fixed structured noise ($\alpha=1$)}}
     \label{fig:fixed_changing}
  \end{subfigure}
 \caption{
 \label{fig:dataset}
 \textbf{Dataset examples.} \textbf{(a)} We introduce distractors to our moving dot dataset by adding either uniform or structured noise with
 different brightness coefficient $\alpha$. \textbf{(b)} Temporally, the noise either changes every frame (top row), or remains fixed throughout the video (bottom row). In the fixed case, the noise is still re-sampled for each new sequence. 
 }
 \vspace{-0.5cm}
\end{figure}

In order to verify whether the proposed JEPA-based methods indeed focus on fixed background noise, we introduce a simple moving dot dataset. The sequences of images contain a single dot on a square with sides of length 1, and the action denotes the delta in the dot's coordinates from the previous time step. We fix the length of the episode to 17 time steps (16 actions).
After pre-training, we probe the representation by training a linear layer to recover the dot's position for all time steps. For pre-training, we use 1 million sequences; for training the prober we use 300,000 sequences; for evaluation, we use 10,000 sequences.
We introduce two types of distractor noise to the background: structured and uniform. We generate structured noise by overlaying CIFAR-10 \citep{cifar10} images. Temporally, the noise can be changing, i.e., each time step the background is resampled; or fixed, i.e., the background does not change with time, but still changes between sequences. For examples, see figure \ref{fig:dataset}. The coefficient $\alpha$ controls noise brightness relative to the dot. We tune hyperparameters of all methods separately for each noise level and type. For more details about the dataset, see appendix \ref{ap:single_dot}.

% \begin{figure}
%   \begin{subfigure}[t]{0.49\linewidth}
%      \centering
%      \includegraphics[width=\textwidth]{figures/structured_fixed.pdf}
%      \caption{\textbf{Structured noise performance}}
%      \label{fig:CFM_Policy}
%   \end{subfigure}
%   \begin{subfigure}[t]{0.49\linewidth}
%      \centering
%      \includegraphics[width=\textwidth]{figures/structured_changing.pdf}
%      \caption{\textbf{Random noise performance}}
%      \label{fig:CFM_Policy}
%  \end{subfigure}
%  \hfill
%  \begin{subfigure}[t]{0.49\linewidth}
%     \centering
%     \includegraphics[width=\textwidth]{figures/uniform_fixed.pdf}
%     \caption{\textbf{Structured noise performance}}
%     \label{fig:CFM_Policy}
%  \end{subfigure}
%  \begin{subfigure}[t]{0.49\linewidth}
%     \centering
%     \includegraphics[width=\textwidth]{figures/uniform_changing.pdf}
%     \caption{\textbf{Random noise performance}}
%     \label{fig:CFM_Policy}
%  \end{subfigure}
%  \hfill
% \end{figure}

\begin{figure}
 \centering
 \includegraphics[width=\textwidth]{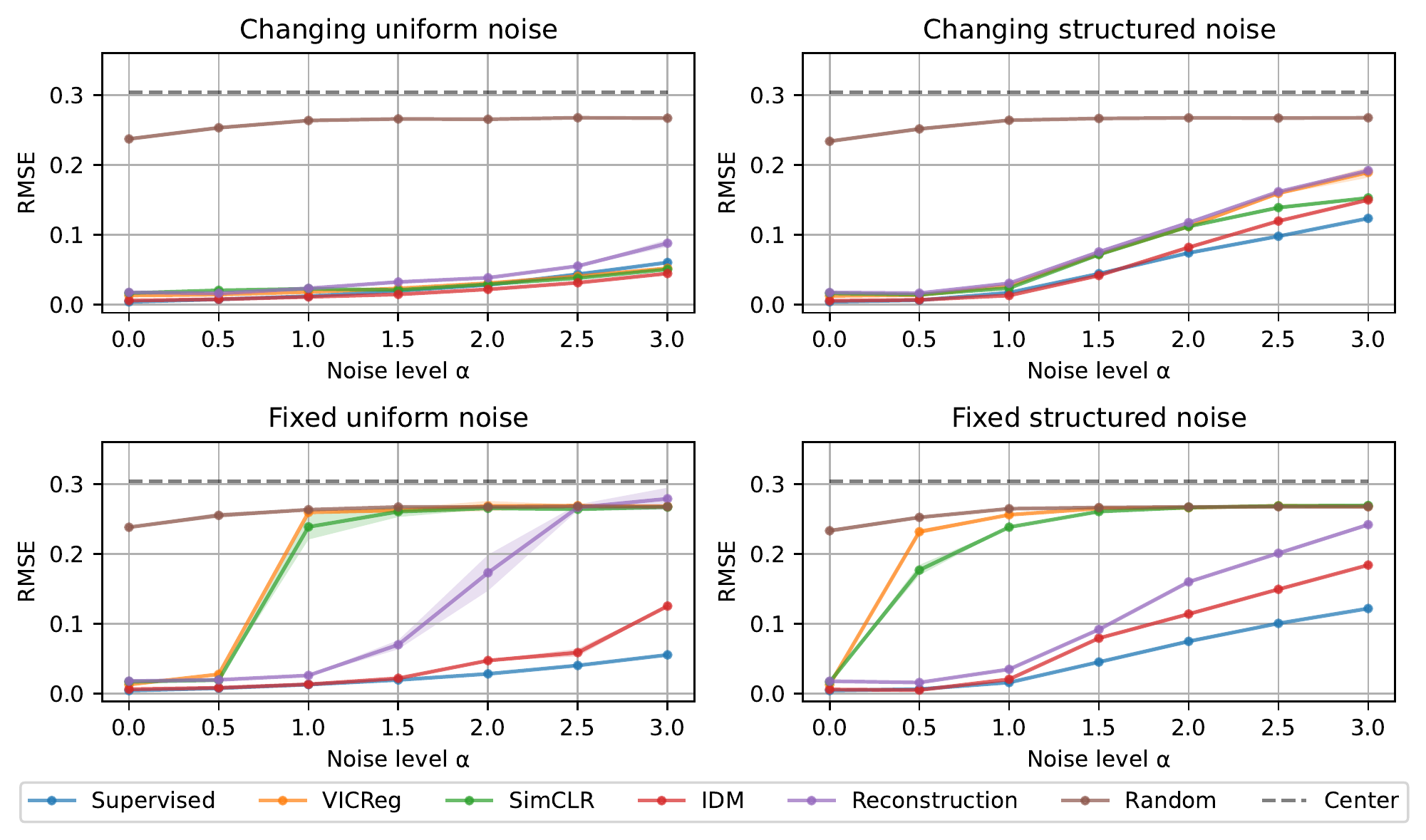}
 \caption{\textbf{Performance of the compared methods with different types and levels of noise.} We tune hyperparameters individually for
 each model, noise level, and type. We show results without tuning in figure \ref{fig:ap:untuned} in the appendix. The dots represent the mean RMSE across 17 time steps. The shaded area represents the standard deviation calculated by running 3 random seeds for each experiment.}
 \label{fig:tuned}
 \vspace{-0.5cm}
\end{figure}

\subsection{Results}

We compare the approaches described in section \ref{sec:method} with different types of noise described above.
We also add a baseline called `Center', which corresponds to always predicting the dot's location to be in the center.
`Center' and `Supervised' baselines should be upper and lower bounds on the error. The results are shown in figure \ref{fig:tuned}.
All methods perform well when there are no distractors. Reconstruction performs well in all settings with $\alpha \leq 1.5$, while JEPA-based methods fail in the presence of fixed noise, both 
structured and uniform. We hypothesize that, as described in section \ref{sec:method}, these methods focus on background noise and
ignore the dot. We observe a similar drop in performance when an extra static dot is introduced instead of distractor noise (see appendix \ref{ap:3_dots_results} for more details). All methods work well when the noise is changing every frame. Additionally, we find that JEPA-based methods do not require hyperparameter tuning to adapt to higher levels of changing noise, while reconstruction performs much worse with untuned hyperparameters (see appendix \ref{ap:additional_results}). Inverse dynamics modeling has great performance in all cases, but this method may be unsuitable for pre-training
as it only learns representations that capture the agent, and fail if there is additional useful information to be captured, as we demonstrate with additional experiments with 3 dots in appendix \ref{ap:3_dots}.

\section{Conclusion}
We demonstrate that JEPA-based methods offer a possible way forward for reconstruction-free forward model learning and are capable of ignoring unpredictable noise well even without additional hyperparameter tuning. However, these methods fail
when slow features are present, even with a large pre-training dataset and hyperparameter tuning. We only demonstrate this with a toy dataset, but we hypothesize that this
may happen in more complex problems. For example, when pre-training a forward-model for self-driving with JEPA on dash-cam videos, the model may focus
on cloud patterns that are easily predictable, rather than trying to learn the traffic participants' behavior. This drawback of JEPA may be addressed by using image differences or optical flow as input to the model, although these input modalities will ignore potentially useful background and may still contain fixed noise. We believe that the way to learn representations that capture both fast and slow features is by adding hierarchy to the architecture (see HJEPA in \cite{lecun2022path}) or by changing the objective to impose an additional constraint that prevents the representations from being constant across time.

\bibliography{main}
\bibliographystyle{abbrvnat}

%%%%%%%%%%%%%%%%%%%%%%%%%%%%%%%%%%%%%%%%%%%%%%%%%%%%%%%%%%%%
\section*{Checklist}

% %%% BEGIN INSTRUCTIONS %%%
% The checklist follows the references.  Please
% read the checklist guidelines carefully for information on how to answer these
% questions.  For each question, change the default \answerTODO{} to \answerYes{},
% \answerNo{}, or \answerNA{}.  You are strongly encouraged to include a {\bf
% justification to your answer}, either by referencing the appropriate section of
% your paper or providing a brief inline description.  For example:
% \begin{itemize}
%   \item Did you include the license to the code and datasets? \answerYes{See Section~\ref{gen_inst}.}
%   \item Did you include the license to the code and datasets? \answerNo{The code and the data are proprietary.}
%   \item Did you include the license to the code and datasets? \answerNA{}
% \end{itemize}
% Please do not modify the questions and only use the provided macros for your
% answers.  Note that the Checklist section does not count towards the page
% limit.  In your paper, please delete this instructions block and only keep the
% Checklist section heading above along with the questions/answers below.
% %%% END INSTRUCTIONS %%%

\begin{enumerate}

\item For all authors...
\begin{enumerate}
  \item Do the main claims made in the abstract and introduction accurately reflect the paper's contributions and scope?
    \answerYes{}
  \item Did you describe the limitations of your work?
    \answerYes{See appendix \ref{ap:limitations}.}
  \item Did you discuss any potential negative societal impacts of your work?
    \answerNA{}
  \item Have you read the ethics review guidelines and ensured that your paper conforms to them?
    \answerYes{}{}
\end{enumerate}

\item If you are including theoretical results...
\begin{enumerate}
  \item Did you state the full set of assumptions of all theoretical results?
    \answerNA{}
        \item Did you include complete proofs of all theoretical results?
    \answerNA{}
\end{enumerate}

\item If you ran experiments...
\begin{enumerate}
  \item Did you include the code, data, and instructions needed to reproduce the main experimental results (either in the supplemental material or as a URL)?
    \answerYes{}
  \item Did you specify all the training details (e.g., data splits, hyperparameters, how they were chosen)?
    \answerYes{}
        \item Did you report error bars (e.g., with respect to the random seed after running experiments multiple times)?
    \answerYes{}
        \item Did you include the total amount of compute and the type of resources used (e.g., type of GPUs, internal cluster, or cloud provider)?
    \answerYes{See appendix \ref{ap:resources}}
\end{enumerate}

\item If you are using existing assets (e.g., code, data, models) or curating/releasing new assets...
\begin{enumerate}
  \item If your work uses existing assets, did you cite the creators?
    \answerYes{See appendix \ref{ap:license}}
  \item Did you mention the license of the assets?
    \answerYes{See appendix \ref{ap:license}}
  \item Did you include any new assets either in the supplemental material or as a URL?
    \answerYes{}
  \item Did you discuss whether and how consent was obtained from people whose data you're using/curating?
    \answerNA{}
  \item Did you discuss whether the data you are using/curating contains personally identifiable information or offensive content?
    \answerNA{}
\end{enumerate}

\item If you used crowdsourcing or conducted research with human subjects...
\begin{enumerate}
  \item Did you include the full text of instructions given to participants and screenshots, if applicable?
    \answerNA{}
  \item Did you describe any potential participant risks, with links to Institutional Review Board (IRB) approvals, if applicable?
    \answerNA{}
  \item Did you include the estimated hourly wage paid to participants and the total amount spent on participant compensation?
    \answerNA{}
\end{enumerate}

\end{enumerate}

%%%%%%%%%%%%%%%%%%%%%%%%%%%%%%%%%%%%%%%%%%%%%%%%%%%%%%%%%%%%

\appendix

\section{Appendix}

\subsection{Related work}

\paragraph{Joint Embedding methods} In recent years, multiple new joint-embedding methods for pre-training image classification systems were introduced \citep{byol, barlow_twins, vicreg, simclr, swav, moco, dim, simsiam}. These methods heavily rely on well-designed image augmentations to prevent collapse \citep{views-contrastive, direct-clr}. 
Very closely related works of \cite{properties, shortcuts} also investigate the tendency of contrastive losses to focus on easy features, although they do not study the application to videos.

\paragraph{Representation learning from video} Many of the ideas from self-supervised pre-training for classification have also been applied to learning from video: \cite{cvrl} applies InfoNCE loss \citep{cpc} to learn encodings of video segments,
while \cite{tcpc} modify CPC architecture \citep{cpc} and include samples from the same video clip as negative examples. \cite{dpc} also propose a modification of CPC \citep{cpc} for training from video and use predicted representations and encoder outputs as positive pair, and construct negative examples using both different videos and different fragments of the same video. \cite{vince} also use videos as an alternative to image augmentations. \cite{tcn} use a triplet loss with single negative example instead of InfoNCE. \cite{video-dim} adopt Deep InfoMax objective \citep{dim} for videos, while \cite{videomoco} use MoCo \citep{moco}. 

\paragraph{Forward model learning} Many methods using reconstruction objectives as main training signal have been proposed \cite{atari_video_fm, dreamer, dreamer_v2, planet}. \cite{dreaming} proposes a reconstruction-free version of Dreamer \citep{dreamer, dreamer_v2} 
using contrastive learning. \cite{bisimulation} proposes a method to learn representations using bisimulation objective, pushing states that
lead to same rewards with same trajectories to have the same representations. In contrast, we focus on the setting where the reward is unknown 
during pretraining. Notably, \cite{bisimulation} also explores the performance of reconstruction and CPC \citep{cpc} objectives in the
presence of distractor noise, but the noise is not fixed in time. 
\cite{offline_rl_test} explore various aspects of model-based and model-free offline RL, and test reconstruction-based RSSM \citep{planet} with various perturbations, concluding that it performs quite well
even with noisy data, which we confirm in our experiments as well.

\paragraph{Self-supervised learning for reinforcement learning} Objectives used in self-supervised learning for images
have been actively used in reinforcement learning from pixels to improve sample complexity of training \cite{spr, curl}, or for pre-training the encoder \cite{sgi, atc, st-dim, light-weight-probing}. These methods focus on pre-training the encoder, and do not evaluate the trained forward model, even if it is present in the system \cite{sgi}.

\subsection{Dataset}
\begin{figure}
     \includegraphics[height=7cm]{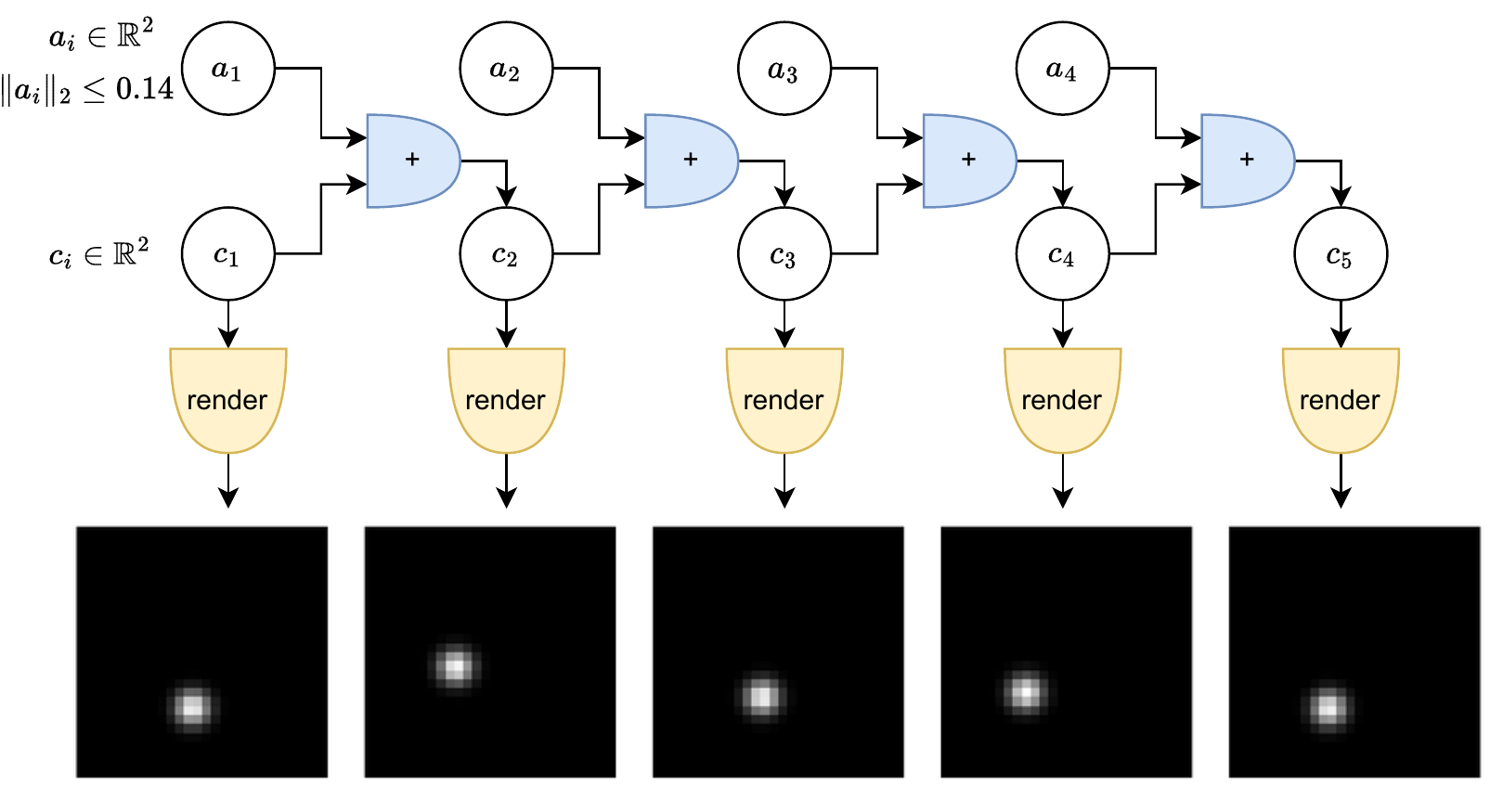}
     \caption{\textbf{Diagram of data generation process.} Initial position $c_1$ and action sequence $a_1 \dots a_T$ are first generated. Then, the actions are sequentially added to the initial location. The resulting coordinates $c_1 \dots c_{T+1}$ are then rendered. }
     \label{fig:ap:data_generation}
\end{figure}
\label{ap:single_dot}
We consider the problem of capturing the location of an object in a video sequence. To this end, we 
introduce a simple environment with just one dot inside a square. 
The dot cannot leave the square, and is always visible on the screen.
We denote dot's coordinates at time $t$ as $c_t = (c_t^x, c_t^y)$.
We assume the square size is 1 by 1, therefore $c_t^x, c_t^y \in [0, 1]$.
At each step, the dot takes an action $a_t = (a_t^x, a_t^y), a_t \in \mathbb{R}^2$.
The norm of $a_t$ is restricted to be less than or equal to a maximum step size $D$: $\Vert a_t \Vert_2 \leq D$.
In our experiments $D=0.14$.

The dot moves continuously around the square by moving by a vector specified by the action $a_t$ with clipping to prevent the 
dot from going outside the square, i.e. $c_{t+1} = \max(0, \min(c_t + a_t, 1))$.

In order to generate one dataset example, we randomly generate a starting location $c_1$, and a sequence of actions $a_{1\dots T}$, where $T$ is the sequence length. In our experiments $T=16$. The actions in a sequence $a_t$ are generated by first sampling vector directions using a Von Mises distribution with a randomly chosen direction $\omega \sim \mathrm{Uniform}(0, 2\pi)$,  $(u^x_t, u^y_t) = u_t \sim \mathrm{VonMises}(\omega, 1.3)$. This prevents the dot from staying in one place in expectation, as would be the case with uniform random sampling. Then we multiply direction vectors with uniformly sampled step size $d_t \sim \mathrm{Uniform}(0, D)$: $a_t = d_tu_t$. We then generate locations by adding the action vectors to the initial location. The length of the action sequence $a_{0\dots T-1}$ is then $T$, while the generated locations sequence is of length $T+1$ (in our case it is 17). A diagram of the process is shown in figure \ref{fig:ap:data_generation}.

Once the actions and locations are generated, we obtain images $o_t$ by rendering the dots at the generated locations $c_t$ applying a Gaussian blur with $\sigma = 0.05$ to the image with the value set to 1 at the location $c_t$. In our experiments, the resolution of $o_t$ is $28 \times 28$.

We introduce distractors to the dataset by overlaying noise onto the dot images.
We consider two noise types: random and structured; and two temporal settings: fixed and changing.
Random noise images $Z$ are images of the same dimension as $o_t$ where each pixel sampled from a uniform distribution, while structured noise images are loaded from CIFAR-10 dataset.

In all cases, we add noise $Z$ with a coefficient $\beta$: $\hat o_t = o_t + \alpha Z_t$. Both noise image and dot image have values between 0 and 1, so the coefficient represents how many times stronger the brightest pixel
in the noise is compared to the brightest pixel in the dot image.

\subsection{Training methods details}
We denote the batch of training video sequences as a tuple of observation and action sequences $(\mathcal{O}, \mathcal{A}), \mathcal{O} = (O_1 \dots O_{T+1}), O_t \in \mathbb{R}^{N \times H \times W} ; \mathcal{A}= (A_1 \dots A_T), A_i \in \mathbb{R}^{N \times M}$. Here, $T$ is the episode length, $H \times W$ is the resolution of observation image, $N$ is batch size, $M$ is the dimension of the action.
For all algorithms we test, the observations are processed using an encoder to obtain representations for each time step $S_i = g_\phi(O_i), S_i \in \mathbb{R}^{N \times d}$, and the forward model is unfolded from the first observation with the given actions: $\tilde S_t = f_\theta(\tilde S_{t-1}, A_{t-1}); \tilde S_1 = S_1$. We use $\mathcal{S}, \mathcal{\tilde S}$ to denote encodings and predictions for all time steps: $\mathcal{S} = (S_1, \dots, S_{T+1})$, $\mathcal{\tilde S} = (\tilde S_1, \dots, \tilde S_{T+1})$. 

\subsubsection{VICReg}
\label{sec:ap_vicreg}
VICReg objective was originally used for image classification \citep{vicreg}. We follow ideas described in \cite{lecun2022path} and
adopt the objective to learning from video. 
We consider representations at each time step separately for calculating variance and covariance losses, while the representation loss becomes the prediction error between forward model and encoder outputs. Total loss and its components are:
\begin{align}
&\mathcal{L}_\mathrm{VICReg} = \alpha \mathcal{L}_\mathrm{prediction} + \beta \mathcal{L}_\mathrm{variance} + \mathcal{L}_\mathrm{covariance} \\
&\mathcal{L}_\mathrm{prediction} = \frac{1}{NT}\sum_{t=1}^T\sum_{i=1}^N\Vert f_\theta(S_{t, i}, A_{t, i}) - g_\phi(O_{t+1, i})\Vert_2^2 = 
\frac{1}{NT}\sum_{t=2}^{T+1}\sum_{i=1}^N \Vert \tilde S_{t, i} - S_{t, i}\Vert_2^2 \\
&\mathrm{Var}(v) = \frac{1}{N-1}\sum_{i=1}^N (v_i - \bar v)^2 \\
&\mathcal{L}_\mathrm{variance} = \frac{1}{T+1}\sum_{t=1}^{T+1}\frac{1}{D}\sum_{j=1}^D \max \left(0, \gamma - \sqrt{\mathrm{Var}(S_{t, :, j}) + \epsilon}\right)  \\
&\mathcal{L}_\mathrm{covariance} = \frac{1}{T+1} \sum_{t=1}^{T+1}\frac{1}{N-1}\sum_{i=1}^D \sum_{j=i+1}^D (S_t S_t^\top)_{i,j} 
\end{align}

\subsubsection{SimCLR}
\label{sec:ap_simclr}
In case of SimCLR adaptation to JEPA, again, we use the same strategy as with VICReg, and treat each time step prediction separately. We
apply SimCLR's InfoNCE loss \citep{simclr} as follows:
\begin{align}
&\mathrm{expsim}(u, v) = \exp \left( \frac{u^\top v}{\tau \Vert u \Vert \Vert v \Vert} \right)\\
&\mathrm{InfoNCE}(S_t, \tilde S_t) = -\frac{1}{N}\sum_{i=1}^{N}\log \frac{\mathrm{expsim}(S_{t, i}, \tilde S_{t, i})}{
\sum_{k=1}^N \mathrm{expsim}(S_{t, i}, \tilde S_{t, i})) + \mathds{1}_{k \neq i} \mathrm{expsim}(S_{t, i}, S_{t, k}) } \\
&\mathcal{L}_\mathrm{SimCLR} = \frac{1}{2T} \sum_{t=2}^{T+1} \mathrm{InfoNCE}(S_t, \tilde S_t) + \mathrm{InfoNCE}(\tilde S_t, S_t)
\end{align}

\subsubsection{Inverse Dynamics Modeling (IDM)}
\label{ap:idm}
We add a linear layer that, given the encoder's outputs at two consecutive
steps $g_\phi(o_t), g_\phi(o_{t+1})$, predicts $a_t$. This should make the encoder pay attention to the parts of
the observation that are affected by the action. The predictor is trained by predicting encoder output. We denote
inverse dynamics model with $h_\xi(s_t, s_{t+1}) = \tilde a_t$. The loss components are then:
\begin{align}
    &\mathcal{L} = \mathcal{L}_\mathrm{prediction} + \mathcal{L}_\mathrm{IDM} \\
    &\mathcal{L}_\mathrm{prediction} = \frac{1}{TN}\sum_{t=1}^T\sum_{i=1}^N\Vert f_\theta(S_{t, i}, A_{t, i}) - g_\phi(O_{t+1, i})\Vert_2^2  \\
    &\mathcal{L}_\mathrm{IDM} = \frac{1}{TN} \sum_{t=1}^T \sum_{i=1}^N \Vert h_\xi(S_{t, i}, S_{t+1, i}) - A_{t, i}\Vert 
\end{align}

\subsubsection{Reconstruction}
\label{ap:rssm}
Reconstruction approach introduces a decoder $d_\xi(\tilde s_t) = \tilde o_t$, and utilizes a reconstruction objective $\mathcal{L} = \frac{1}{T}\sum_{t=1}^{T} \Vert o_t - \tilde o_t \Vert_2^2$ to train the encoder and predictor. The decoder mimics the architecture of the encoder, and uses
up-sampling followed by convolutional layers to match strided convolutional layers in the encoder. The total loss is:
\begin{align}
    &\mathcal{L}_\mathrm{reconstruction} = \frac{1}{T}\sum_{t=2}^{T+1} \Vert d_\xi(\tilde s_t) - o_t \Vert^2_2\\
\end{align}

% RSSM uses sequences of $\{ o_t, a_t\}_{t=1}^{T}$. Here $t$ is discrete timestep, $o_t$ is image observation. $a_t$ is action vector. (this is explained in the beginning of the section)
% RSSM is a pixel reconstrution based method. It separates the state $s_t$ into deterministic and stochastic parts: $s_t = (s_t^\mathrm{det}, s_t^\mathrm{stoch})$, the predictor then only predicts the deterministic part, while the prior and posterior models sample the stochastic part. 
% Deterministic state model predicts the deterministic part of the state: $\tilde{s}_t^\mathrm{det} = f_\theta(\tilde{s}_{t-1}^\mathrm{det}, s_{t-1}^\mathrm{stoch}, a_{t-1})$. 
% Stochastic state model learns a distribution of stochastic states given the deterministic state: $\tilde{s}_t^\mathrm{stoch}  \sim p(\tilde{s}_t^\mathrm{stoch} | \tilde{s}_t^\mathrm{det})$. We also train a posterior model that has access to the last observation: $ q( s_t ^ \mathrm{stoch} | \tilde{s}_t^\mathrm{det}, s_t)$. We train a decoder with the architecture that mimics the encoder to represent $p(o_t | s_t)$. Then, the total loss is:
% \begin{align}
% \mathcal{L}_\mathrm{RSSM} &= \mathcal{L}_\mathrm{reconstruction} + \mathcal{L}_\mathrm{KL-divergence} \nonumber\\
% &= \sum_{t=1}^{T}\left( \mathbb{E}_{q(\tilde{s}_t|o_{\le t})} [\ln p(o_t | \tilde{s}_t)]) - \mathbb{E}[KL[ q(s_t ^ \mathrm{stoch} | \tilde{s}_t^\mathrm{det}, s_t)\vert\vert p(\tilde{s}_t^\mathrm{stoch} | \tilde{s}_t^\mathrm{det})
%  ]] \right) \nonumber\\
% \end{align}

\subsection{Probing details}
\label{ap:probing}
In order to test whether representations contain the desired information, we train a probing function $q(s) : \mathbb{R}^D \to \mathbb{R}^Q$ which we represent with one linear layer. $D$ is the representation size, $Q$ is the target value dimension. In our case, the target value is the dot's location, so $Q=2$. When training the prober,
we follow a similar protocol to the pre-training. Given the initial observation $o_1$ and a sequence of actions of length $T$, $a_1 \dots a_T$, we encode the first observation $\tilde s_1 = g_\phi(o_1)$,
then auto-regressively apply the forward model $\tilde s_t = f_\theta(\tilde s_{t-1}, a_{t-1})$, obtaining representations for a sequence of length $T+1$: $\tilde s_1 \dots \tilde s_{T+1}$. We denote the location of the dot in the $i$-th batch element at time step $t$ as $C_{t, i}$. Then, the loss function we apply to train $q$ is:
\begin{align}
\mathcal{L}_\mathrm{probing} = \frac{1}{TN}\sum_{t=1}^{T}\sum_{i=1}^N \Vert q(\tilde S_{t,i}) - C_{t, i} \Vert_2^2
\end{align}

\subsection{Additional results}
\label{ap:additional_results}
\begin{figure}
 \centering
 \includegraphics[width=\textwidth]{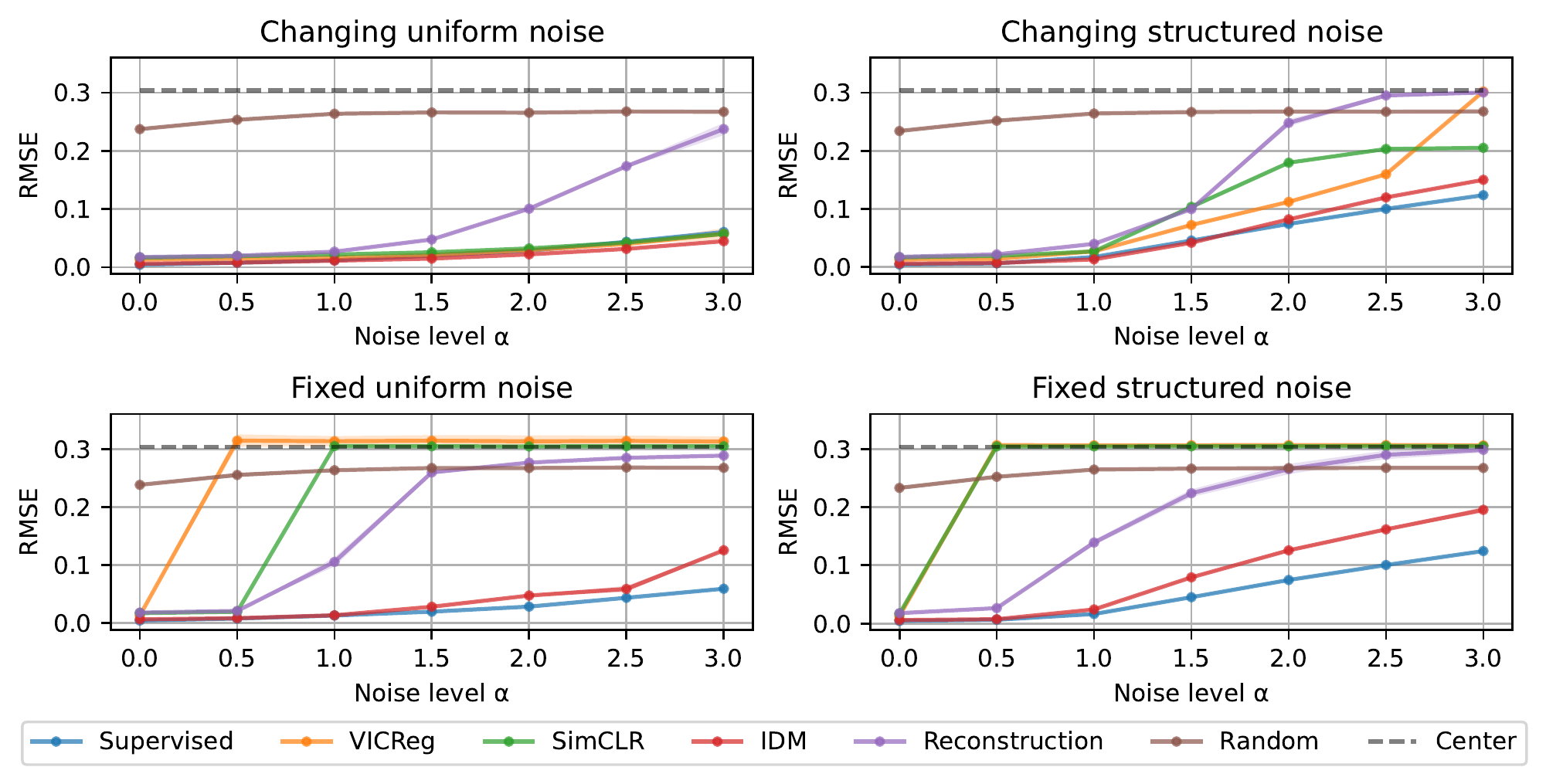}
 \caption{\textbf{Performance of compared methods with different types and levels of noise.} Hyperparameters for each model are chosen for no-noise setting and, in contrast to figure \ref{fig:tuned} are not re-tuned for each type or level of noise.
 The dots represent the mean RMSE across 17 time steps. The shaded area shows standard deviation calculated by running 3 random seeds for each experiment.}
 \label{fig:ap:untuned}
 \vspace{-0.5cm}
\end{figure}

In figure \ref{fig:ap:untuned} we show results for the same experiments as in figure \ref{fig:tuned}, with one difference that we do not tune hyperparameters for each model and noise setting. We pick the best hyperparameters for noise-free setting, and
fix them for all levels of noise. We see that SimCLR and VICReg JEPA methods, compared to tuned performance, fail with lower levels of noise, while reconstruction fails with changing uniform noise without tuning.

\subsection{3 dots dataset}
\label{ap:3_dots}

\subsubsection{Environment}
\begin{figure}
\begin{center}
     \includegraphics{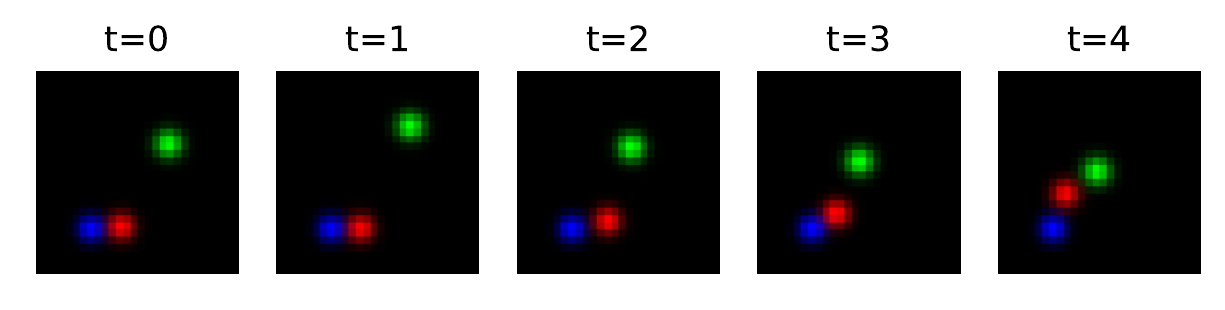}
     \caption{\textbf{Example sequence of 3 dots dataset.} Red, the first channel, corresponds to the action controlled dot. Green corresponds to the randomly moving dot with unknown actions. Blue corresponds to the stationary dot.}
     \label{fig:ap:3dots_env}
\end{center}
\end{figure}

In this dataset, there are three dots on the three channels of the image: action-controlled dot, uncontrollable dot, and stationary dot. The first dot is action controlled similar to the single dot environment without any noise. The second dot is similar to the first dot, but the actions are unknown, making it impossible to predict its movements. The third dot is stationary across all the frames of the episode, but is at different positions in different episodes. To generate a sample of the three dot dataset, we concatenate these dots along different channels of image. These fixed channels ensure that our model can distinguish between the dots. We show an example sequence in figure \ref{fig:ap:3dots_env}.
The goal is to learn a representation that captures the locations of the action-controlled and stationary dots, while ignoring the random dot.

\subsubsection{Results}
\label{ap:3_dots_results}
\begin{table}
\caption{\textbf{Results for 3-dots dataset.} All numbers denote RMSE across 17 steps. We run 3 seeds for each experiment to obtain standard deviations. Cells are colored according to the values, with higher values shown in red, and lower values shown in blue.}
\begin{center}
\begin{tabular}{ccccc}
Method & Average & Action & Random & Stationary \\
\hline \\
VICReg&\cellcolor{red!22.900000000000002}0.229±0.031&\cellcolor{red!27.700000000000003}0.277±0.041&\cellcolor{red!27.3}0.273±0.044&\cellcolor{blue!10.200000000000001}0.066±0.026\\SimCLR&\cellcolor{red!15.8}0.158±0.001&\cellcolor{red!19.3}0.193±0.001&\cellcolor{red!19.3}0.193±0.002&\cellcolor{blue!22.500000000000004}0.025±0.001\\IDM&\cellcolor{red!23.400000000000002}0.234±0.001&\cellcolor{blue!19.5}0.035±0.000&\cellcolor{red!29.799999999999997}0.298±0.002&\cellcolor{red!27.200000000000003}0.272±0.000\\Supervised&\cellcolor{red!10.4}0.104±0.000&\cellcolor{blue!27.000000000000004}0.010±0.001&\cellcolor{red!18.0}0.180±0.001&\cellcolor{blue!28.5}0.005±0.000\\Reconstruction&\cellcolor{red!10.7}0.107±0.000&\cellcolor{blue!23.7}0.021±0.001&\cellcolor{red!18.2}0.182±0.000&\cellcolor{blue!22.200000000000003}0.026±0.002\\Random&\cellcolor{red!26.0}0.260±0.001&\cellcolor{red!23.5}0.235±0.002&\cellcolor{red!27.800000000000004}0.278±0.002&\cellcolor{red!26.5}0.265±0.004\\Center&\cellcolor{red!29.9}0.299±0.000&\cellcolor{red!30.4}0.304±0.000&\cellcolor{red!30.4}0.304±0.000&\cellcolor{red!28.9}0.289±0.000\

\end{tabular} 
\end{center}
\end{table}

We show performance of the compared methods on 3-dots dataset in table
\ref{ap:3_dots_results}. Hypeparameters were tuned for each model to
obtain the best average RMSE. VICReg and SimCLR based methods focus only on the stationary dot, and fail to capture the other two dots.
Again, we hypothesize that JEPA methods capture `slow features' \citep{slow_feature_analysis}, and with the stationary dot containing the slowest features, the other dots are ignored. IDM manages to capture the action-controlled dot, but ignores the stationary dot, as it is
irrelevant to inverse dynamics. Reconstruction-based approach captures both the stationary and action-controlled dots.

\subsection{Model architectures}
\label{ap:architectures}

\textbf{The encoder} consists of 3 convolutional layers with ReLU and BatchNorm after each layer, and average pooling with kernel size of 2 by 2 and stride of 2 at the end. The first convolution layer has kernel size 5, stride 2, padding 2, and output dimension of 32. The second layer is the same as the first, except the output dimension is 64. The final layer has kernel size 3, stride 1, padding 1, and output dimension of 64. After average pooling, a linear layer is applied, with output dimension of 512.

\textbf{The predictor} is represented by a single-layer GRU \citep{gru} with hidden representation size of 512, and input size of 2. Hidden representation is initialized at the first step of predictions with encoder output $g_\phi(o_1)$. The inputs at each time step are actions.

\textbf{Reconstruction} introduces a decoder represented by a model symmetric to the encoder, with convolutions in reverse order and upsampling. We do not use a latent variable in our implementation.

\subsection{Limitations}
\label{ap:limitations}
The limitation of the current experiments is the use of the simple toy dataset. It is unclear that the same will hold for more complicated video datasets and bigger models. Another limitation is that we only check VICReg and SimCLR losses for JEPA methods, while there are many more objectives, e.g., \cite{byol, barlow_twins}.

\subsection{Computational resources}
\label{ap:resources}
All experiments were run using AMD MI50 or Nvidia RTX 8000 GPUs. For each noise type and level, 100 random hyperparameters were run, and the best one was run for 3 seeds. Each individual experiment takes less than one hour of GPU time.

\subsection{Code license}
\label{ap:license}
To implement reconstruction-based approach, we used parts of the implementation of \cite{RSSM_code} distributed under MIT license. Implementing InfoNCE loss, we used the code from \cite{SimCLR_code}, which is also distributed under MIT license.

\subsection{Acknowledgements}
This material is based upon work supported by the National Science Foundation under NSF Award 1922658.

\end{document}